\documentclass[runningheads]{llncs}
\usepackage{url}

\usepackage{float}
\usepackage{times}
\usepackage{latexsym}
\usepackage{lingmacros}
\usepackage[whole]{bxcjkjatype}
\usepackage{gb4ea}
\usepackage{color}
\usepackage{amsmath}
\usepackage{multirow}
\usepackage{graphicx}
\usepackage{here}
\usepackage{tikz,subfigure}
\usetikzlibrary{automata,positioning,arrows,shapes,backgrounds,trees}
\usepackage{tikz-qtree}
\usepackage{tikz-qtree-compat}
\usepackage{stmaryrd}
\usepackage {natbib} 
\newcommand{\refex}[1]{(\ref{ex:#1})}

\newcommand{\refdef}[1]{Definition~\ref{def:#1}}

\newcommand{\refth}[1]{Theorem~\ref{th:#1}}

\newcommand{\reflem}[1]{Lemma~\ref{lem:#1}}

\newenvironment{centre}{\begin{center}$\begin{array}{c}}{\end{array}$\end{center}}
\usepackage{fancybox}
\usepackage{dashbox}
\newenvironment{basicshadowbox}%
  {\medskip%
   \noindent%
   \begin{Sbox}%
   \begin{minipage}{.95\textwidth}%
   }{%
   \end{minipage}%
   \end{Sbox}%
   \shadowbox{\TheSbox}} 
\newenvironment{basicstackedbox}%
  {\medskip%
   \noindent%
   \begin{Sbox}%
   \begin{minipage}{.95\textwidth}%
   }{%
   \end{minipage}%
   \end{Sbox}%
   \lbox[1]{\fbox{\TheSbox}}
   }
  {\medskip%
   \noindent%
   \begin{Sbox}%
   \begin{minipage}{.95\textwidth}%
   }{
   \end{minipage}%
   \end{Sbox}%
   \dbox{\TheSbox}
   }
\cornersize{.2}
  {\medskip%
   \noindent%
   \begin{Sbox}%
   \begin{minipage}{.95\textwidth}%
   }{
   \end{minipage}%
   \end{Sbox}%
   \ovalbox{\TheSbox}
   }
  {\resetcounter{equation}%
   \begin{basicshadowbox}%
   \begin{definition}[#1]%
   }{%
   \end{definition}%
   \end{basicshadowbox}} 
  {\resetcounter{equation}%
   \begin{basicshadowbox}%
   \begin{axiom}[#1]%
   }{%
   \end{axiom}%
   \end{basicshadowbox}}
  {\resetcounter{equation}%
   \begin{basicstackedbox}%
   \begin{theorem}[#1]%
   }{
   \end{theorem}%
   \end{basicstackedbox}%
   }
  {\resetcounter{equation}%
   \begin{basicstackedbox}%
   \begin{lemma}[#1]%
   }{
   \end{lemma}%
   \end{basicstackedbox}%
   }
  {\resetcounter{equation}%
   \begin{basicshadowbox}%
   \noindent\textbf{\refdef{#2}}\ #1%
   }{%
   \end{basicshadowbox}%
   } 
  {\resetcounter{equation}%
   \begin{basicshadowbox}%
   \begin{axiom}[#1]%
   }{%
   \end{axiom}%
   \end{basicshadowbox}}
  {\resetcounter{equation}%
   \begin{basicstackedbox}%
   \noindent\textbf{\refth{#2}}\ #1%
   }{
   \end{basicstackedbox}%
   }
  {\resetcounter{equation}%
   \begin{basicstackedbox}%
   \noindent\textbf{\reflem{#2}}\ #1%
   }{
   \end{basicstackedbox}%
   }

\newcommand{\resetcounter}[1]{\setcounter{#1}{0}}

\usepackage{logic}
\usepackage{proof}
\usepackage{ccg}
\usepackage{dts}
\usepackage{caption}
\begin{document}
\mainmatter              
\title{Matrix and Relative Weak Crossover \\
in Japanese: \\
An Experimental Investigation
}
\titlerunning{Matrix and Relative Weak Crossover in Japanese}
%
%
\author{Haruka Fukushima\inst{1} \and Daniel Plesniak\inst{2} \and Daisuke Bekki\inst{1}}
\authorrunning{Fukushima, Plesniak, and Bekki} 
%
\tocauthor{Haruka Fukushima, Daniel Plesniak, and Daisuke Bekki}
\institute{Ochanomizu University, Tokyo, Japan\\
\email{\{fukushima.haruka, bekki\}@is.ocha.ac.jp}
\and
Seoul National University, Seoul, Republic of Korea\\
\email{plesniak@usc.edu}}

\maketitle              

\begin{abstract}
This paper provides evidence that weak crossover effects differ in nature between matrix and relative clauses. \cite{Fukushima2024} provided similar evidence, showing that, when various non-structural factors were eliminated English speakers never accepted matrix weak crossover cases, but often accepted relative weak crossover ones. 
Those results were limited, however, by English word order, which lead to uncertainty as to whether this difference was due to the effects of linear precedence or syntactic structure. 
In this paper, to distinguish between these two possibilities, we conduct an experiment using Japanese, which lacks the word-order confound that English had. 
We find results that are qualitatively in line with \cite{Fukushima2024} suggesting that the relevant distinction is structural and not based simply on precedence.
\end{abstract}
\section{Introduction}
\label{sec:introduction}
Under a Chomskyan theory of language, a, or perhaps the, fundamental property of the human language faculty is its capacity to generate recursive structures (\cite{Hauser2002}). 
A classic manifestation of this is the embedding of clauses within other clauses. 
While such embedding can be direct, with clauses as ``arguments'' of the verbs in higher clauses, e.g., [$_{clause}$ John said [$_{clause}$ Mary thinks [$_{clause}$ \dots]]], it can also be indirect, with embedded clauses occurring within other elements themselves embedded in the clause. Such is the case of relative clauses, where the clauses are embedded in nominal phrases within higher clauses, e.g., [$_{clause}$ John met [$_{nominal}$ a woman [$_{clause}$ who knows [$_{nominal}$ a man [$_{clause}$ who \dots]]]]].  
Such clauses frequently appear to differ more radically in form from the matrix (highest) clause than their directly embedded counterparts do, as in the English examples provided in this paragraph, where the relative clauses are (a) marked with a wh-element and (b) are ``missing'' the logical subject of the verb, which instead appears outside the clause: in matrix `a man praised a child' becomes in relative `who praised a child', with `man' being the noun ``modified'' by the relative clause (which we will term its ``head''). 
The properties of these clauses, particularly the relationship between the clause and its head, in comparison and/or contrast with matrix clauses and their arguments, is thus of pressing interest for understanding the human linguistic faculty.
	
 To narrow the focus slightly, consider that, while languages differ in which elements of a relative clause may be ``missing'', the two most common such elements are subjects and objects (\cite{KeenanandComrie1977}). 
 In matrix (and directly embedded clauses), it has long been argued that subjects and objects display asymmetric properties, because, at least on a Chomskyan view, they occupy different structural positions, with the object forming a constituent (the predicate) with the verb to the exclusion of the subject, such that the subject is structurally ``higher'' than the object  (e.g., as formalized by \cite{Reinhart1976} in terms of c-command). 
 The existence of such a structural asymmetry, itself evidence for a recursive, step-by-step, structure-building operation, such as \cite{Chomsky1995}'s Merge, is supported in part by the existence of a priori unexpected interpretative constraints. 
 To cite one crucial example, as argued in \citet{Reinhart1983}, bound variable anaphora (BVA) interpretations, such as (1) are frequently said to be possible if the ``binder'' is the subject and the ``bindee'' is in the object, but not if this situation is reversed, as in (2). 
 Further, such a constraint cannot be reduced to a matter of surface word order nor to issues of simple conceptual/semantic constraints such as thematic role of the binder and bindee, as shown by the reported acceptability of sentences like (3) and (4) respectively.
\begin{exe}
    \ex \label{1} Every boy praised his teacher.\\
    BVA=each boy praised his own teacher, John praised John's, Bill Bill's, etc.\\
Typical judgement: BVA possible
    \ex \label{2} His teacher praised every boy\\
    BVA=each boy was praised by his own teacher, John's praised John, Bill's Bill, etc.\\
Typical judgement: BVA impossible
    \ex \label{3} His teacher, every boy praised\\
    BVA=same as (\ref{1})\\
Typical judgement: BVA possible
    \ex \label{4} Every boy was praised by his teacher.\\
BVA=same as (\ref{2})\\
Typical judgement: BVA possible
\end{exe}

Following~\cite{Wasow1972}, we may term the somewhat surprising unacceptability of (\ref{2}) a ``weak-crossover'' effect. 
 While there are many possible theories that can account for the existence of such effects, the robust detection of analogous patterns of judgement across languages imposes constraints on the possible theories of human linguistic competence, particularly as concerns the generation and interpretation of matrix clauses and their directly embedded counterparts. 
 And robust indeed its seems to be; while~\cite{Wasow1972} correctly noted the ``weakness'' of the constraint, as expressed through both intra-and inter-speaker variation on the (in)acceptability of BVA interpretations in such configurations, such variation can be accounted for under an enriched theory of BVA, such as that proposed by \cite{Ueyama1998}. 
 In a nutshell,~\cite{Ueyama1998}'s theory proposes that BVA interpretations can derive from multiple sources, only one of which is strictly conditioned on structure. 
 Other sources are conditioned on word-order or semantic/pragmatic factors.
 \cite{Hoji2022a,Hoji2022b,Hoji2022c}, supported by further experimental evidence from \cite{Plesniak2022a,Plesniak2022b,Plesniak2023}, shows that, once non-structural factors are controlled for, judgement variation on weak-crossover sentences vanishes, and they are universally rejected, even while the other types of sentences discussed above continue to be accepted. 
 The existence of weak-crossover effects may thus be regarded as a clear manifestation of the subject-object asymmetry in matrix clauses.
	
 The natural question is whether such effects extend to relative clauses as well, i.e., whether there is such a thing as ``relative weak-crossover'' (henceforth R-WCO) analogous to ``matrix WCO'' (henceforth M-WCO).
 To illustrate, the relative-clause equivalents of (\ref{1}) and (\ref{2}) would be as in (\ref{5}) and (\ref{6}), with (\ref{6}) being the R-WCO case:
\begin{exe}
    \ex \label{5} Every boy who praised his teacher (was rewarded.)
    \ex \label{6} Every boy who his teacher praised (was rewarded.)
\end{exe}
 
  Depending on one's theory of relative clauses and the interpretation thereof, we may or may not expect there to be such an effect. 
  Unlike with M-WCO, there has been a distinct lack of unanimity in the literature in terms of what the correct judgements on R-WCO cases are, and thus, what sorts of theories should be proposed. 
  In proposals such as many of those overviewed in \cite{Safir2017}, it is assumed that the ``missing'' relevant clause subject/object position is occupied by either (a) (a copy/trace of) the head itself, (b) (a copy/trace of) the wh-element, which ``corresponds'' to the head in the relevant sense, or (c) some other ``corresponding'' element; in all cases, it is this element that enters into the BVA relationship with the pronominal element, and thus, if it is in the object position, a weak-crossover configuration is created.
	
 On the other hand, other hypotheses do not force an element in the ``missing'' object position to be the ``binder''. 
One relevant example is the analysis of WCO in \cite{Bekki2021} using Dependent Type Semantics (DTS).
DTS assumes combinatory categorial grammar (CCG: \citet{Steedman1996,Steedman2000}) as the syntactic theory and uses underspecified dependent type theory (UDTT) as the semantic theory.
Semantic composition is a homomorphic map from a syntactic structure of CCG to a preterm of UDTT. 
However, anaphoric expressions and presupposition triggers in a syntactic structure are mapped to the special type of UDTT called underspecified types. 
The underspecified type requires the given type to have a proof in the formation rule, and the construction of this proof corresponds linguistically to anaphora resolution and presuppositions filtering.
In other words, in DTS, anaphora accessibility is proof constructability in a given environment, which also explains the WCO effects.
In the semantic representations of (\ref{5}) and (\ref{6}), the occurrences of \textit{his} introduce underspecified types, but in both (\ref{5}) and (\ref{6}), the underspecified types are included in the restriction scope of \textit{every}.
Thus, in both cases, the underspecified type can refer to a universally quantified variable introduced by \textit{every}, allowing for a proof construction using it.
Therefore, a BVA reading is expected to be present in both cases (\ref{5}) and (\ref{6}).
This differs from cases (\ref{1}) and (\ref{2}): In (\ref{1}), the underspecified type appears within \textit{every}'s nuclear scope, and therefore, proof can be constructed using the universally quantified variable introduced by \textit{every} (and thus the presence of BVA reading is predicted).
On the other hand, in (\ref{2}), the underspecified type is not in any of \textit{every}'s restriction/nuclear scopes, and therefore, proof construction that produces BVA readings is not possible.
	
 The goal of the present study is to distinguish between the predictions of these two different families of theories. 
 In other words, this study investigates whether purported R-WCO effects display the same basic empirical pattern of behavior as their M-WCO counterparts.
 Our basic finding is that they do not; BVA interpretations of purported R-WCO configurations are frequently accepted. 
 Though, as mentioned above, it is true that BVA interpretations of M-WCO configurations are also sometimes accepted, we can clearly distinguish between these two cases. 
 In the M-WCO cases, \cite{Hoji2022b}'s diagnostics allow us to accurately predict BVA acceptance, while with R-WCO, the data defy such predictions. 
 Interpretatively, this suggests that, while M-WCO acceptance is permitted only due to the influence certain specific non-structural sources, R-WCO can occur without the need of such sources.
 As such, the only reasonable conclusion available at this time is that R-WCO can arise due to structural sources, supporting hypotheses of the \cite{Bekki2021}-like family of hypotheses, i.e., those which predict asymmetries between matrix and relative clauses.

\section{Previous investigations}
\label{sec:previousworks}
\cite{Fukushima2024} conducted an experiment in English that stands as the direct predecessor to our current one. 
It in turn drew on experiments previously mentioned above, such as \cite{Hoji2022c} and \cite{Plesniak2022a,Plesniak2022b,Plesniak2023}, utilizing the methodology laid out in \cite{Hoji2015,Hoji2022a}'s ``Language Faculty Science'' paradigm. 
With minor modifications, our experimental design and analysis follows that of \cite{Fukushima2024} and thus, we roughly summarize the experiment from that study here.
In the experiment, participants were shown sentences alongside with image that displayed interpretations. 
A typical example was as in (\ref{7}) below:
\begin{exe}
    \ex  \label{7} Example BVA item from \cite{Fukushima2024}\\
    ``More than one school in Japan lied to its high-achieving student''
\begin{figure}
    \centering
    \includegraphics[width=6cm]{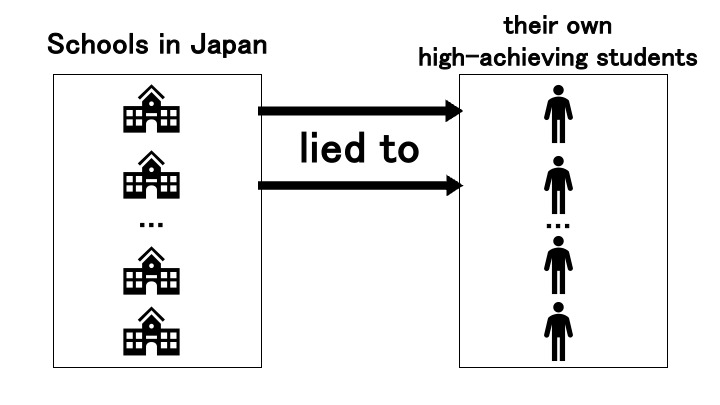}
    \includegraphics[width=6cm]{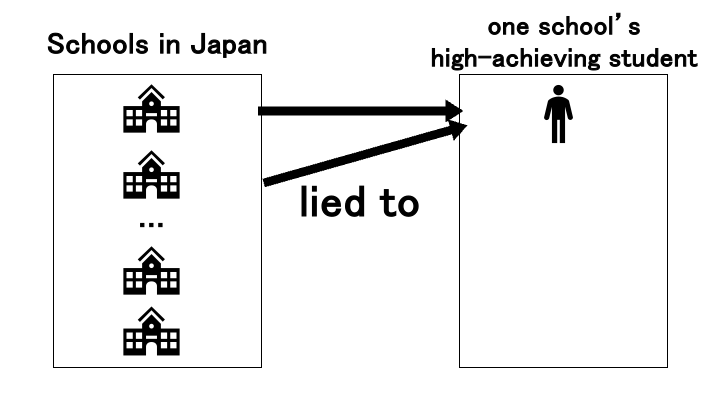}
    \caption{Accompanying images of sentence (\ref{7})}
    \label{fig:figure1}
\end{figure}
\end{exe}

Participants were asked to select which of the pictures could represent possible interpretations of the sentence. Selecting both options or neither was also possible. 
As can be noted, in (\ref{7}) the option on the lefthand side represents a BVA interpretation of the sentence (where `more than one school in Japan' binds `its'), whereas the option on the right represents a non-BVA interpretation, where `its' is interpreted according to a fixed referent.
This design was general to all items in the experiment: a sentence accompanied by two images, one consisting of a target interpretation and another of a non-target interpretation. 
Selecting the equivalents of either ``both'' or ``just the target'' was taken as indicating the individual's acceptance of the BVA (or other, see below) interpretation for the sentence in question.
In addition to checking the acceptability of the main items of interest, e.g., BVA interpretations of M-WCO and R-WCO configurations, several other types of items were included for the purposes of noise control, following the general model of \cite{Hoji2015,Hoji2022c}. 
The first of these were so-called ``sub-experiments'', testing each participant's attentiveness to, comprehension of, and general compatibility with the experiment. 
To unpack this concept briefly, a participant's reported judgements might not match their actual judgements if (a) they are not paying attention to the task and/or are answering randomly, (b) they do not understand the questions as intended, or (c) they are being clever/resourceful in a way unanticipated or undesired by the experimenter. \footnote[1]{To give a clear example of this last possibility, consider the sentence `every boy praised John's mother.' It is widely reported that names cannot be understood as bound variables, and as such `every boy' cannot ``bind'' `John' in such a sentence, i.e., it cannot mean something like `each boy praised his own mother'. Yet as one of our authors has observed, undergraduate students, when confronted with such sentences, occasionally insist that such a reading is possible, their logic invariably being some version of: ``well, maybe each boy is named John''. The students are neither being inattentive nor failing to understand what type of interpretation is being addressed. They are, however, being  ``too clever'' for the task at hand, i.e., exploiting a sort of meta-linguistic understanding of the phrase ``John's mother'' alongside imagining a highly unusual situation where everyone is named John. While an interesting observation, we suspect few would take such judgements to invalidate the generalization that names cannot be construed as bound variables. As such, they constitute a form of ``noise'' for the purposes of assessing the validity of generalizations about BVA acceptability, and thus should be controlled for.}
We will describe the analogous sub-experiments of the current experiment in the following section, so for the sake of space, we give just one example of \cite{Fukushima2024}'s sub-experiments here:
\begin{exe}
    \ex Sub-experiment example from \cite{Fukushima2024}\\
    \label{8} ``Every school in Japan lied to the same high-achieving student.''
\begin{figure}
    \centering
    \includegraphics[width=6cm]{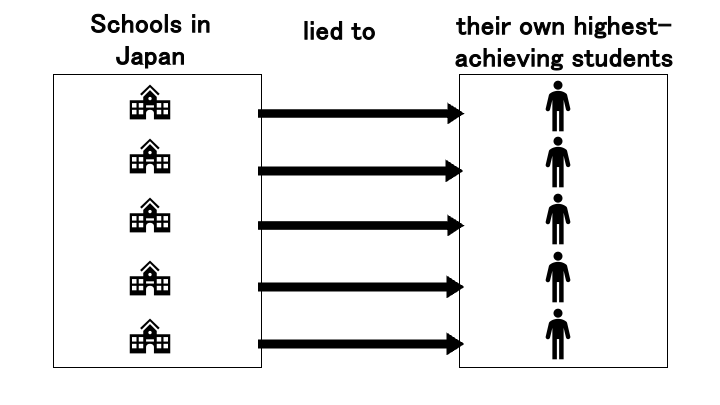}
    \includegraphics[width=6cm]{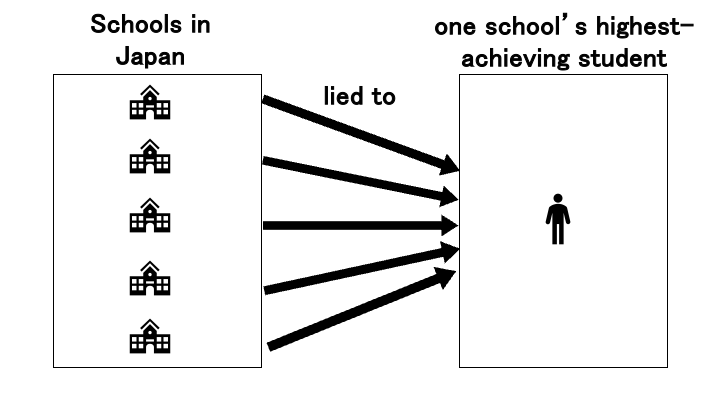}
    \caption{Accompanying images of sentence (\ref{8})}
    \label{fig:figure2}
\end{figure}
\end{exe}
On the assumption that no individual who is paying attention, understands the question being asked, and is not making use of some ``excessively clever'' maneuver will agree that `the same high-achieving student' ought to refer to one individual in this case, i.e., cannot be bound. 
As a result, such a person ought not to say either that `both' pictures or `just the one on the left' can be valid interpretations of the sentence. 
If someone does say so, then we have principled grounds to classify their responses on the crucial sentences of interest as ``not fully reliable'' for hypotheses-testing purposes. 
Only those whose answers to sub-experiment questions of this type are consistently ``accurate'' can be considered to have fully significant judgements for the purposes of evaluating the success or failure of a given prediction. 
Even if a participant's judgements are accurately reported, however, it does not necessarily mean they have direct bearing on every prediction.
This is because, as discussed in the previous section, according to models of BVA acceptability such as that of~\cite{Ueyama1998}, there are different sources of BVA, only one of which directly reflects structural relationships. 
As such, any structure-based prediction about BVA acceptability will robustly hold only when no other sources of BVA are available.
It is thus necessary to determine whether other sources are available for a given individual at a given time in order to evaluate whether structure-based predictions are expected to hold for them.

In particular, \cite{Ueyama1998} predicts that, in the case of weak-crossover, BVA may be possible via the use of ``quirky binding''. 
Following \cite{Hoji2022b}, we understand quirky binding to be the reflex of certain semantic-pragmatic ways of achieving BVA (or at least BVA-like) readings, which \cite{Hoji2022b} terms NFS1 and NFS2 (NFS standing for ``non-formal source''). 
\cite{Hoji2022a,Hoji2022b} presents a method for detecting whether NFS1 or NFS2 is possible for a given speaker (at a given time) on a given sentence type, which, as it turns out, seems to be heavily dependent on the choice of binder and bindee. 
\cite{Hoji2022a}'s solution is to examine the individual's pattern of acceptance for other types of readings which involve, separately, the binder or the bindee. 
In particular, to assess the status of the bindee, distributive reading (DR) interpretations are used, and for the binder, coreferential (Coref) readings.
(\ref{9}) and (\ref{10}) below are examples of these in \cite{Fukushima2024}'s experiment, respectively.

\begin{exe}
    \ex DR \\
    \label{9}``More than one school in Japan lied to two high-achieving students.''
\begin{figure}[h]
    \centering
    \includegraphics[width=6cm]{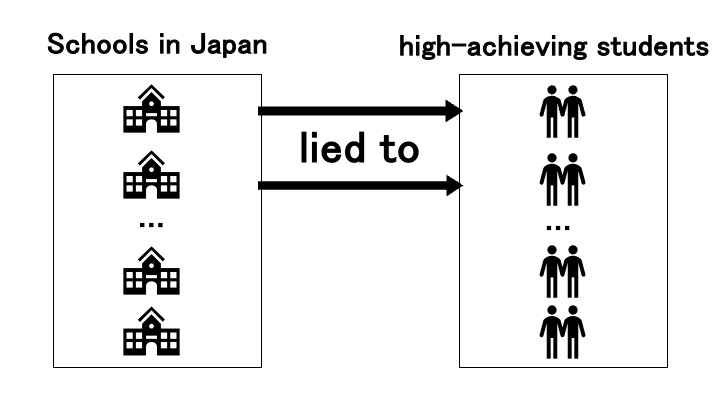}
    \includegraphics[width=6cm]{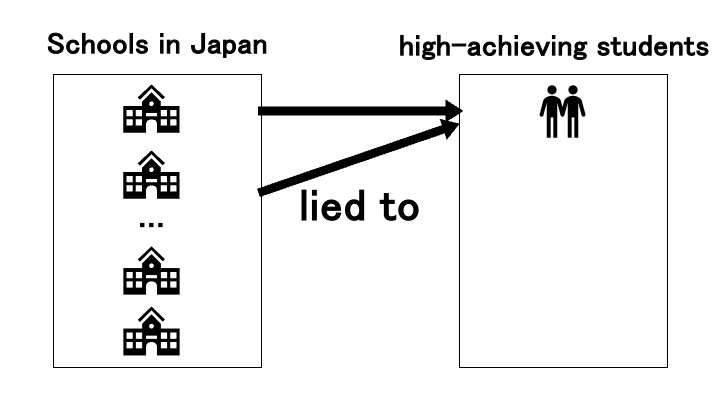}
    \caption{Accompanying images of sentence (\ref{9})}
    \label{fig:figure3}
\end{figure}
    \ex Coref \\
    \label{10}``A school in Japan lied to its high-achieving student.''
\begin{figure}[h]
    \centering
    \includegraphics[width=6cm]{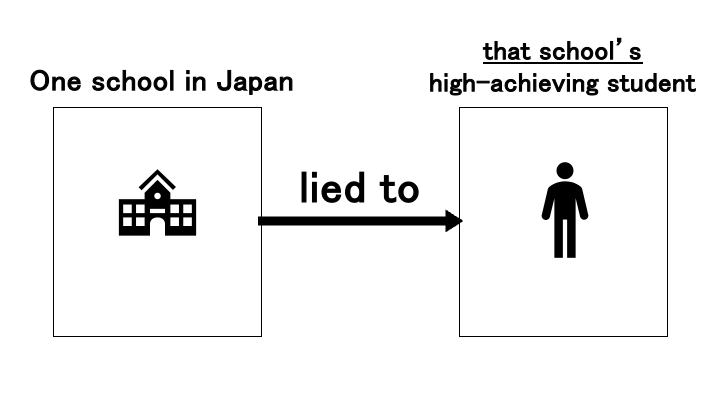}
    \includegraphics[width=6cm]{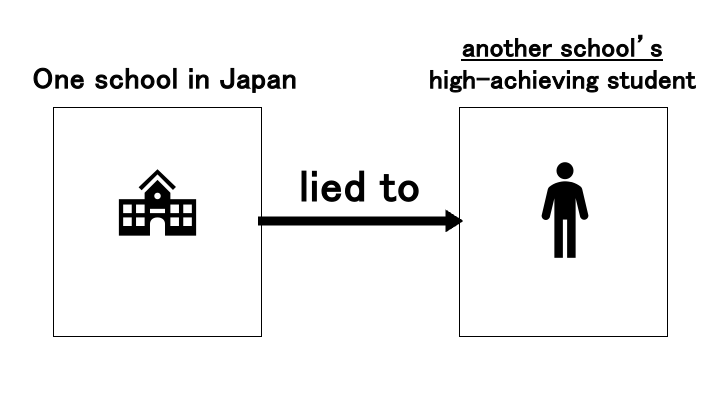}
    \caption{Accompanying images of sentence (\ref{10})}
    \label{fig:figure4}
\end{figure}
\end{exe}
In the DR interpretation (the left picture in (\ref{9})), the `two high-achieving students' in question vary based on the school, as opposed to the non-DR interpretation (on the right) where `two high-achieving students' refers to a specific two students. 
In the Coref interpretation, (the left picture in (\ref{10})), `it' refers to the same school as `a school' in the sentence, whereas in the non-Coref interpretation (on the right), `it' refers to a different school. 

Note that these two interpretations involve either the binder `more than one school in Japan' or bindee `it' of items like (\ref{7}), but crucially (a) only one of them, and (b) considered with regard to an interpretation other than BVA. 
By examining an individual's pattern of judgements about the acceptability of DR/Coref with the binder/bindee with regard to weak-crossover analogous such as (\ref{11}) and (\ref{12}), it is possible to diagnose the presence/absence of NFS1/NFS2.

\begin{exe}
    \ex \label{11}Two high-achieving students lied to more than one school in Japan.
    \ex \label{12}Its high achieving student lied to a school in Japan.
\end{exe}
We describe the exact process of this analysis later in this paper, so we will delay details until then; \cite{Fukushima2024}'s analytic process is substantively analogous. 
What is crucial is that, after controlling for attentiveness and other issues with sub-experiments, as well as the potential for ``quirky binding'' with DR and Coref, \cite{Fukushima2024} replicate previous findings with regard to M-WCO: of all individuals meeting all the diagnostic criteria in order for their reported judgements to be considered significant relative to the predictions, not one of them accepts BVA with a M-WCO configuration. 
R-WCO sentences, on the other hand, do not follow this pattern; they are nearly universally accepted with BVA readings, and attempting to control for NFS1/NFS2 via DR and Coref does little to change this. 
As such, \cite{Fukushima2024} conclude that M-WCO and purported R-WCO configurations are asymmetric with regard to their BVA possibilities, as predicted by accounts like \cite{Bekki2023}.

However, as \cite{Fukushima2024} admit, their results suffer from a serious confounding factor, namely that, in contrast to M-WCO, R-WCO cases in English all involve the binder preceding the bindee, as can be seen in sentences like (\ref{2}) and (\ref{6}), repeated as (\ref{13}) and (\ref{14}) below.
\begin{exe}
    \ex \label{13}His teacher praised every boy.
    \ex \label{14}Every boy who his teacher praised (was rewarded.)
\end{exe} 
Given that, in the \cite{Ueyama1998} model of BVA, binders preceding bindees can potentially enable non-structural source of BVA, it cannot be definitively concluded that the M-WCO R-WCO contrast observed is due to structure; it could simply be due to differing linear word order. 
As such, current experiment seeks to distinguish between these two possibilities, utilizing Japanese, where relative clauses precede their head nouns, and as such, the word order in M-WCO and purported R-WCO sentences is matched.
As noted in the previous section, what we find is that the distinction persists. 
Therefore, while we cannot rule out an influence of a precedence-based source on \cite{Fukushima2024}'s English results, we conclude that such results cannot be based exclusively on such a source, and thus, that structure does indeed play a role.

\section{Experimental design}
\label{sec:designofexperiment}
Our experiment largely follows the design of \cite{Fukushima2024} discussed above. 
The main BVA sentence types are as below:
\begin{table}[H]
    \centering
    \begin{tabular}{l l l l l}
    \hline
    \textbf{Sentence Type} & \textbf{Abbreviation} &\textbf{Form} &\textbf{X c-commands Y } & \textbf{X precedes Y }\\
    \hline
    \hline
    Canonical BVA & SOV & X-ga Y-no N-o/ni V& yes & yes \\
    Matrix Weak Crossover & WCO & Y-no N-ga X-o/ni V.& no & no\\
    Major Subject & MS &  X-ga Y-no N-ga V & yes & yes \\
    Reconstruction & OSV & Y-no N-o/ni X-ga V& yes & no \\
    Subject Relative Clause & SRC & Y-no N-o/ni $V_1$ X-ga $V_2$& no & yes \\
    Object Relative Clause & ORC & Y-no N-ga $V_1$ X-ga $V_2$& ? & no\\
    \hline
    \end{tabular}
    \caption{C-command relation and precedence in the main BVA sentence type}
    \label{tab:table1}
\end{table}

 SOV and WCO instantiate the basic (matrix) weak crossover contrast discussed in the introduction, i.e., contrast cases where the subject is the subject and the bindee is in the object (SOV) and vice versa (WCO). 
 OSV is a scrambled version of SOV. 
 Its use is to demonstrate that the SOV vs. WCO rejection cannot simply be ascribed to matters of linear word order; in both WCO and OSV, bindee precedes binder, but by hypothesis, OSV sentences allow for reconstruction, enabling the scrambled object to be interpreted as an object in an SOV sentence, i.e., meeting the conditions for structure-based BVA. 
 SRC (subject relative clause) and ORC (object relative clause) are the matrix clause equivalents of SOV and WCO respectively. 
 In terms of the preceding discussion, our ``WCO'' corresponds to ``M-WCO'' and ``ORC'' corresponds to ``R-WCO''; however, as the existence or non-existence of an ``R-WCO'' effect is precisely the question we are attempting to address, we give the corresponding sentence pattern the more theory-neutral label of ORC for now, restricting WCO to the, already established, M-WCO. 
 Finally, MS is a ``major subject'' construction 
(\cite{Kuno1973}, \cite{Mikami1960}) , which we use to check one additional possible hypothesis as to the nature of relative clause structure, to be discussed in the conclusion section.
	
 Each BVA sentence pattern is instantiated by two sentences, and each of these sentences is itself matched with two DR and Coref analogues, in order to effect the detection of NFS-type BVA sources, as discussed in the previous section. One example of each type of item (BVA, DR, and Coref) is given below, each instantiating an SOV construction.\footnote[2]{In the experiment, sentences were provided in Japanese, but we present them via the standard linguistic gloss here.
}
\clearpage
 \begin{exe}
     \ex \label{15} BVA
     \gll  Jidoosya-gaisya-3sya-ga soko-no syatyoo-o hihanshita. \\
     Three-car-companies-NOM it-GEN president-ACC criticized.\\
     \trans`Three car companies criticized their presidents.'
    \begin{figure}[ht]
        \centering
        \includegraphics[width=1.0\linewidth]{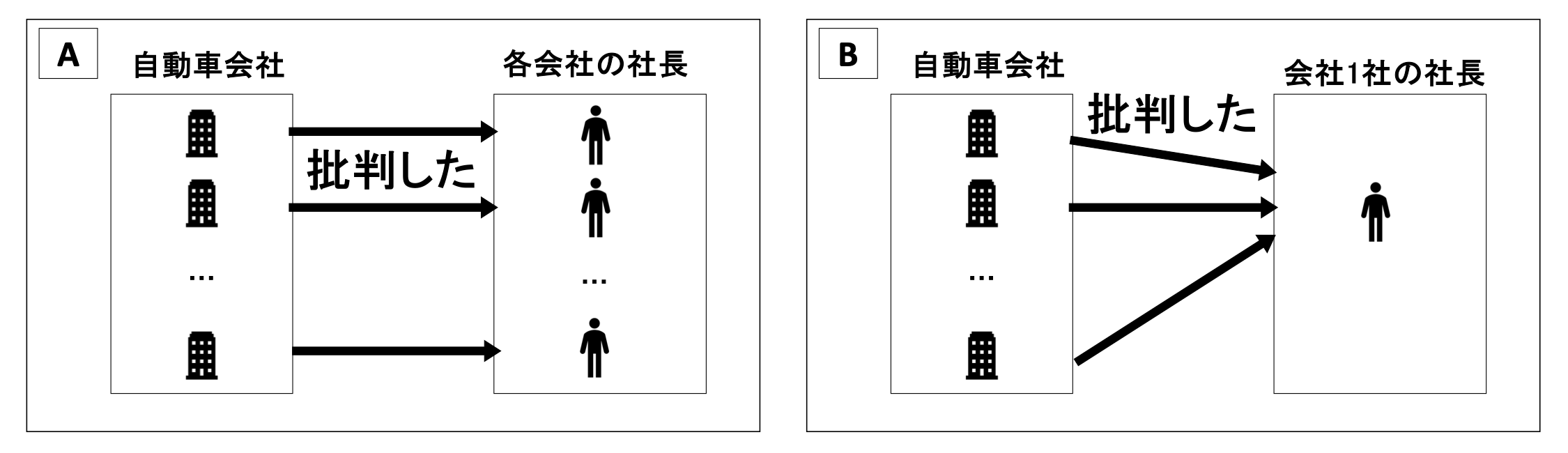}
        \caption{Example pictures of (\ref{15})}
        \label{fig:figure5}
    \end{figure}
     \ex \label{16}DR
     \gll Jidoosya-gaisya-3sya-ga butyoo-2ri-o hihanshita.\\
     Three-car-companies-NOM two-managers-ACC criticized.\\
     \trans `Three car companies criticized two managers.'
    \begin{figure}[ht]
        \centering
        \includegraphics[width=1.0\linewidth]{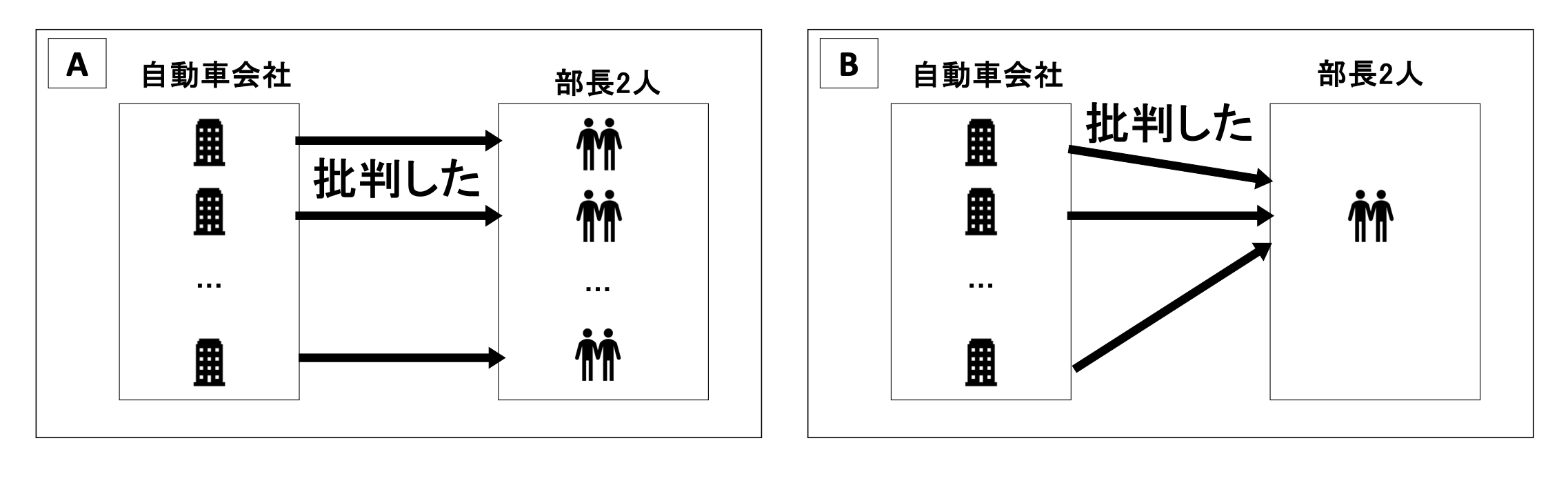}
        \caption{Example pictures of (\ref{16})}
        \label{fig:figure6}
    \end{figure}
    \ex \label{17}Coref 
     \gll Aru-jidoosya-gaisya-ga soko-no syatyoo-o hihanshita. \\
      A-car-company-NOM  it-GEN president-ACC criticized.\\ 
      \glt `A car companies criticized its presidents.'
    \begin{figure}[ht]
        \centering
        \includegraphics[width=1.0\linewidth]{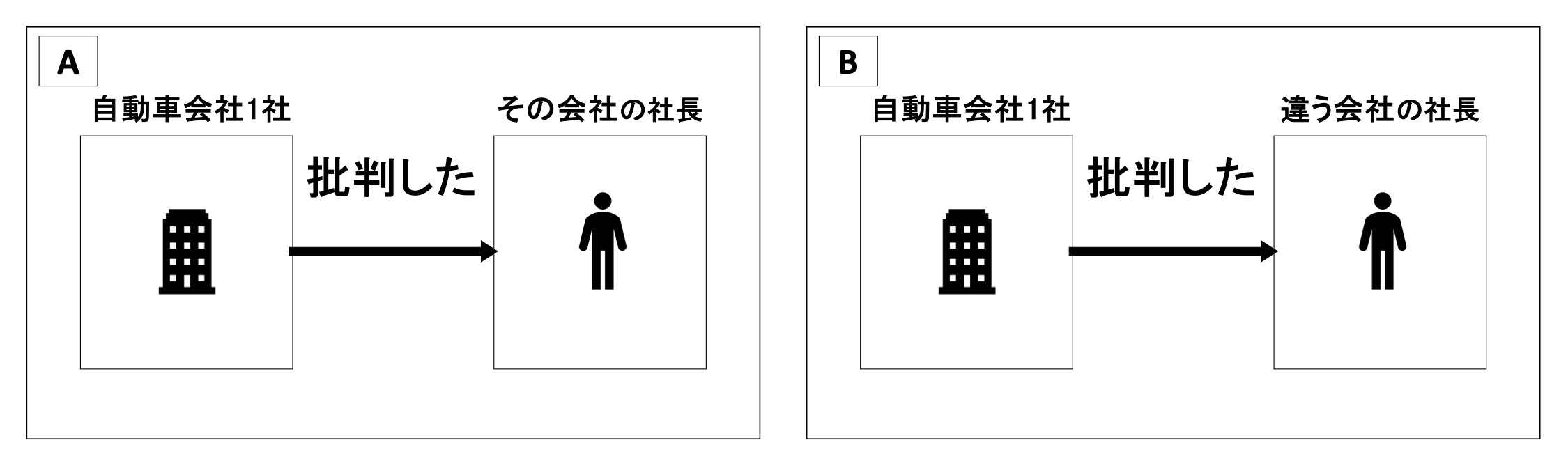}
        \caption{Example pictures of (\ref{17})}
        \label{fig:figure7}
    \end{figure}
 \end{exe}

We also include sub-experiments as described in the previous section. 
They are:
\begin{exe}
    \ex Instruction Sub-experiments (12 total, 4 per each of BVA, DR, and Coref)
    \gll \label{18}Subete-no jidoosya-gaisya-ga ano-syatyoo-o hihansita.\\
    Every-GEN car-company-NOM that-president-ACC criticized.\\
     \glt`Every car company criticizes that president.'
    \begin{figure}[ht]
        \centering
        \includegraphics[width=1.0\linewidth]{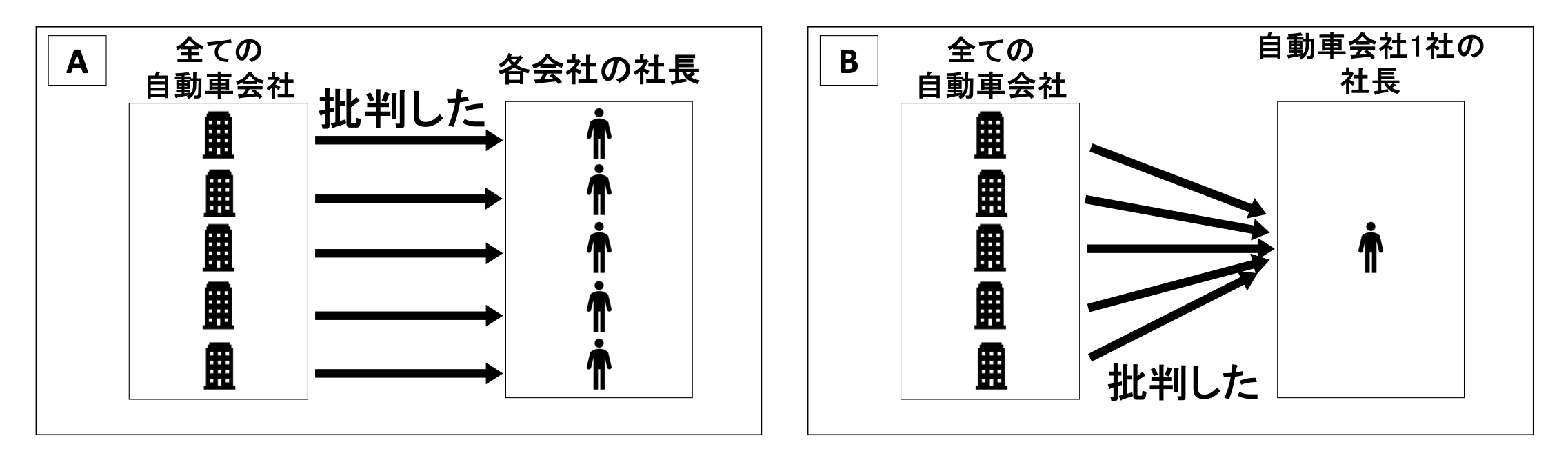}
        \caption{Example pictures of (\ref{18})}
        \label{fig:figure8}
\end{figure}
\ex Lexical Sub-experiments (12 total, 4 per each of BVA, DR, and Coref )
\gll \label{19}Subete-no jidoosya-gaisya-ga asoko-no-syatyoo-o hihansita.\\
    Every-GEN car-company-NOM over-there-GEN-president-ACC criticized.\\
    \trans `Every car company criticizes the president over there.'
\begin{figure}[ht]
        \centering
        \includegraphics[width=1.0\linewidth]{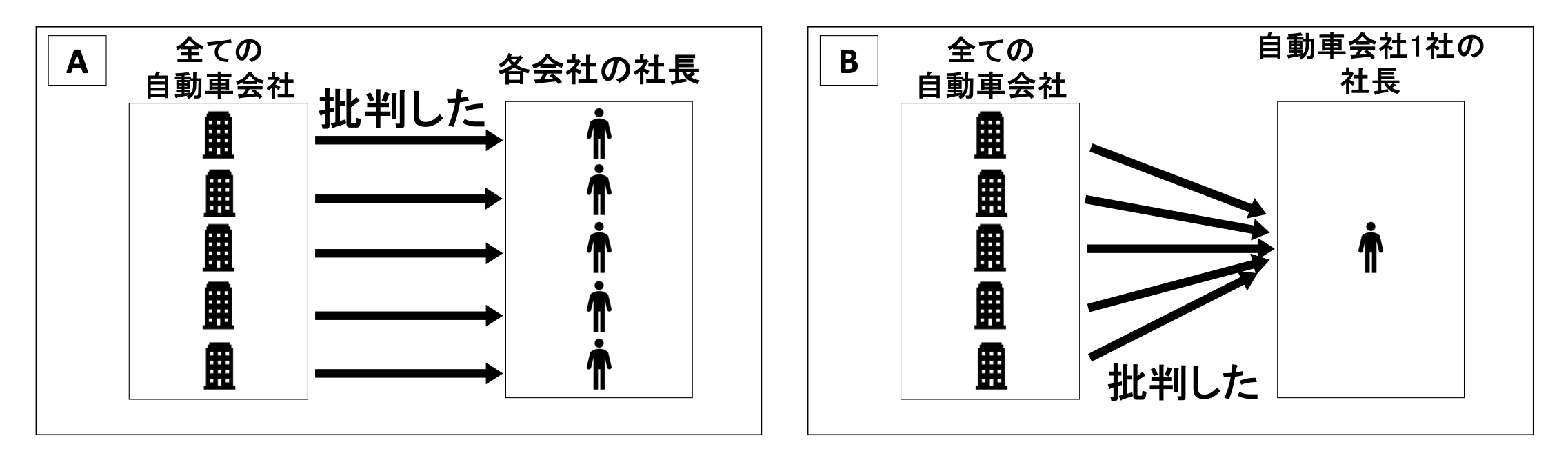}
        \caption{Example pictures of(\ref{19})}
        \label{fig:figure9}
\end{figure}
\clearpage
\ex \label{20} Topic Sub-experiments (8 total)
\gll Asoko-no syatyoo-o, aru-jidoosya-gaisya-ga Toyota-no butyoo-o hiannsita.\\
Over-there-GEN president-ACC, a-car-company-NOM Toyota-GEN manager-ACC criticized.\\
 \trans `The president over there, a car company criticized Toyota's manager.'
 \begin{figure}[ht]
        \centering
        \includegraphics[width=1.0\linewidth]{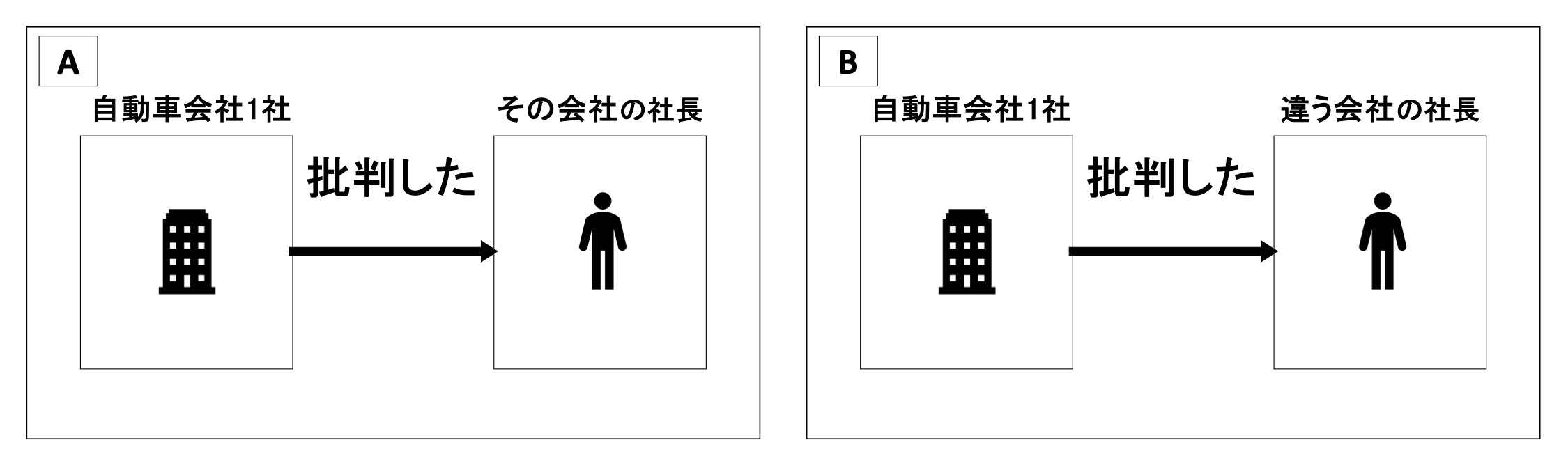}
        \caption{Example pictures of (\ref{20})}
        \label{fig:figure10}
    \end{figure}
\ex \label{21} Split Coreference Sub-experiments (4 total)
\gll Toyota-ga Nissan-o soko-no syatyoo-ni-kansuru-kennde hihansita.\\
 Toyota-NOM Nissan-ACC it-GEN president-about criticized.\\
\trans `Toyota criticized Nissan about its president.'
 \begin{figure}[ht]
        \centering
        \includegraphics[width=1.0\linewidth]{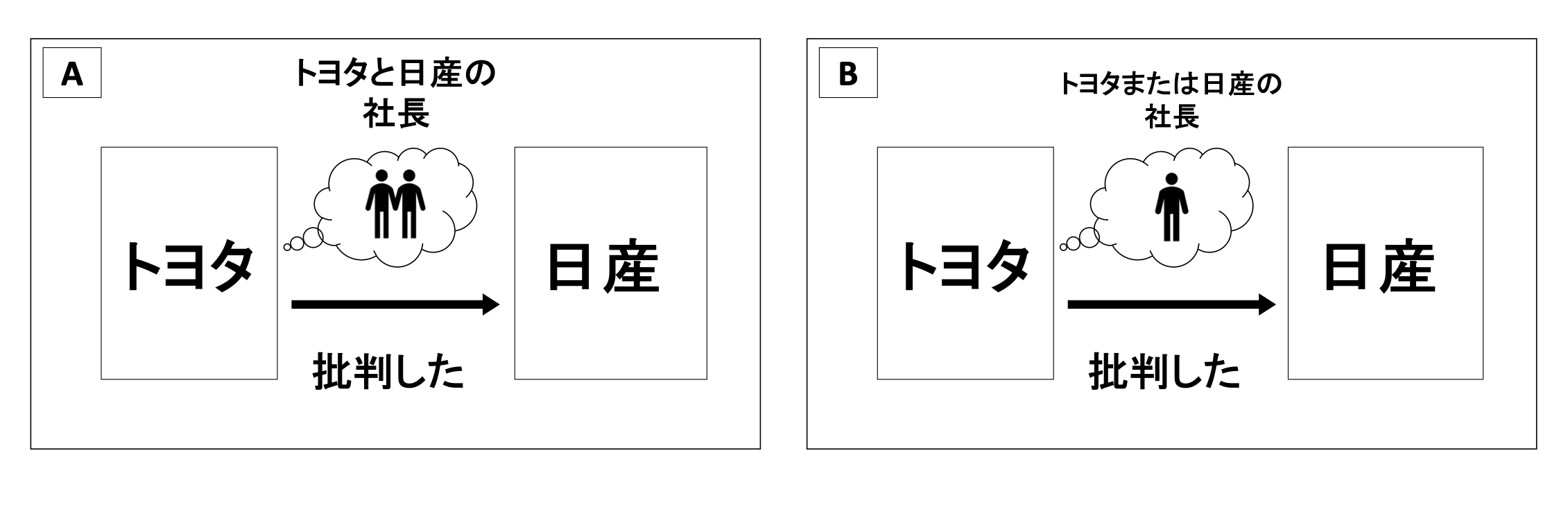}
        \caption{Example pictures of (\ref{21})}
        \label{fig:figure11}
    \end{figure}   
\end{exe}

For reasons of space, we refer readers to \cite{Hoji2015} for a full discussion of the nature of these sub-experiments. 
 In brief, first, Instruction Sub-experiments are tests to check whether informants understand the interpretations of sentences and pictures.
 In the case of (\ref{18}), picture A of Fig.\ref{fig:figure8} has a BVA-style interpretation, and picture B has an interpretation where there is just one `president' involved. 
 Since the sentence refers to \textit{ano president} `that president', only the latter interpretation should be possible, so an informant indicating that picture A is a possible interpretation indicates a potential problem. 
 Likewise,the Lexical Sub-experiments function analogously, testing whether an informant reports accepting BVA/DR/Coref readings with elements that are not supposed to be able to yield such readings, such as BVA with \textit{asoko} `that place' in (\ref{19}) .
 Topic Sub-experiments test whether informants understand scrambled sentences in the intended way, i.e., they do not somehow take the scrambled object to refer to someone different than the semantic object of the verb. 
 In the pictures of (\ref{20}), one picture has exactly that kind of interpretation, and thus should be rejected by attentive participants. 
 Finally, in the case of the Split Sub-experiments, as in (\ref{21}), we test whether participants can accept elements like \textit{soko} `that place' as jointly coreferring to two distinct entities at the same time, as can English `their' in sentences like `John and Bill spoke about their joint project.' 
 This would be undesirable (as well as contradict typically reported Japanese judgements), as it would allow for coreferntial readings that are effectively indistinguishable from BVA readings, making it unclear which sort of interpretation a participant's reported judgements are about. Thus, picture A of Fig.\ref{fig:figure11} has a split coreferential reading, picture B a non-split one, and we thus check whether participants disallow the split reading as intended.

As it turns out, in our analysis to be presented below, we find that we do not need to make crucial use of these sub-experiments, so, for the purpose of the particular results to be described, the sub-experiments can also be considered filler items.

\section{Results}
\label{sec:results}
Even before any detailed analysis is performed, we can see clear contrasts in acceptability tendencies between the WCO sentence pattern (M-WCO) and the ORC sentence pattern (R-WCO) simply by looking the rates of acceptance:
    \begin{figure}[h]
        \centering
        \includegraphics[width=0.8\linewidth]{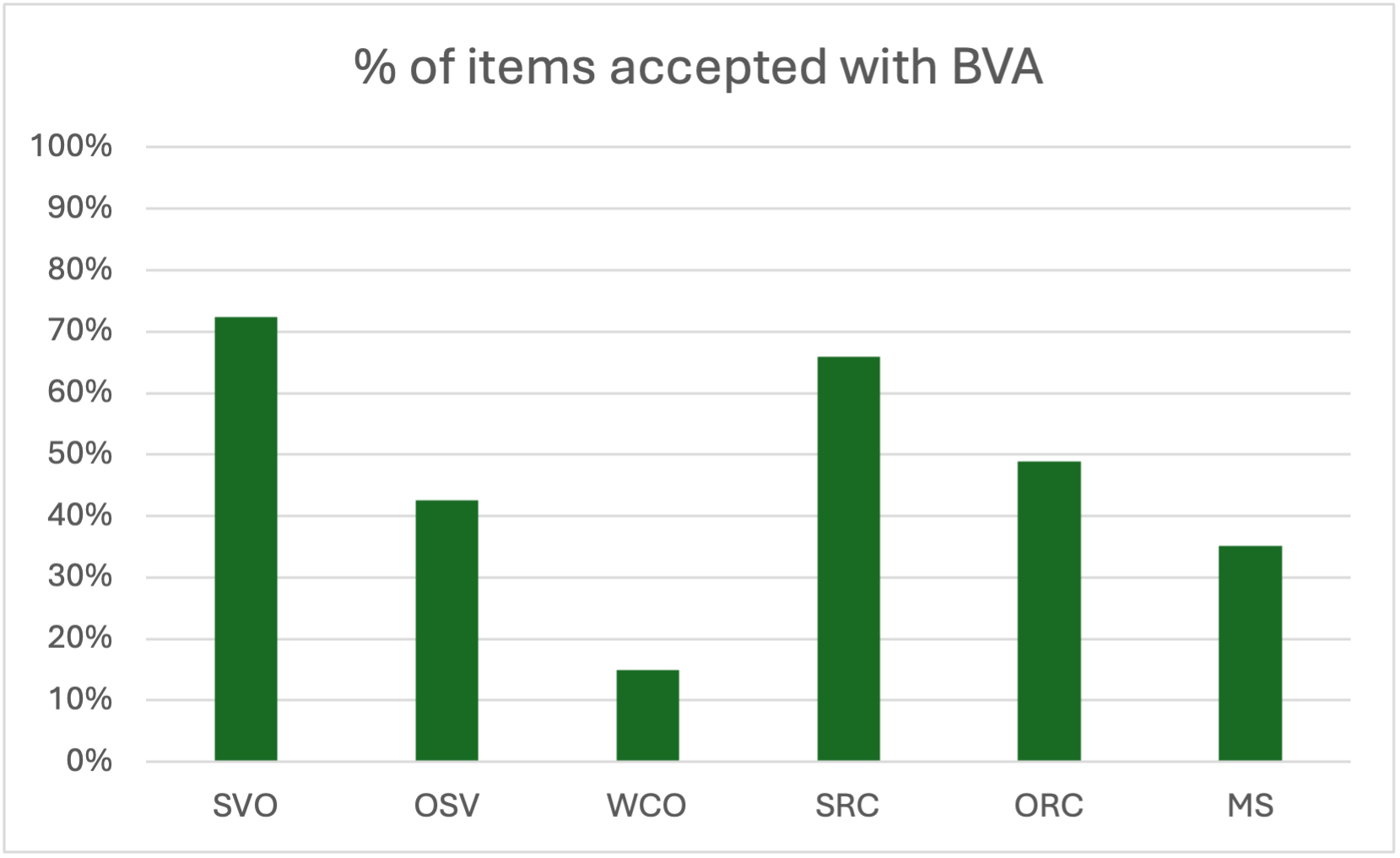}
        \caption{Percentage of sentences accepted with BVA}
        \label{fig:figure12}
    \end{figure}
    \clearpage
    \begin{figure}[h]
        \centering
        \includegraphics[width=0.8\linewidth]{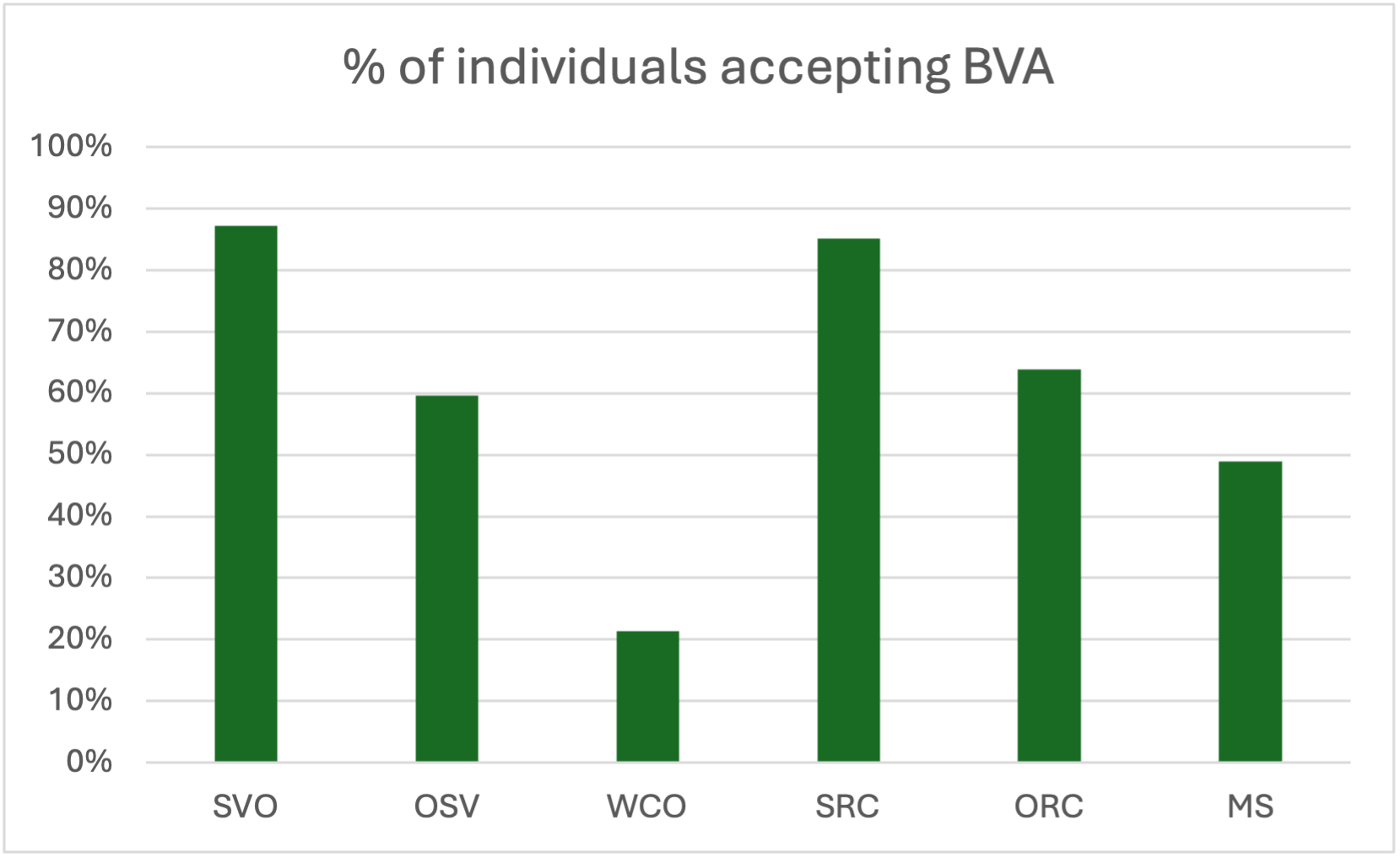}
        \caption{Percentage of individuals accepting any example of a sentence type with BVA}
        \label{fig:figure13}
    \end{figure}
While the rate of acceptance of BVA in ORC sentences is not as high as that of the canonical SOV sentences, note that it, in both cases above, is similar to and indeed higher than the acceptance rate of BVA in OSV sentences. 
That structure-based BVA is acceptable in Japanese OSV (reconstruction) sentences is not particularly controversial (see the references cited above, e.g., \cite{Ueyama1998}); that they are not accepted as frequently as their SOV counterparts can be attributed (at least in part) to the fact that BVA where binder follows bindee is often disprefered, both for pragmatic reasons and due to the unavailability of the precedence-based source. 
Once this ``penalty'' is taken into account, the rate of ORC acceptance is consistent with expectations for a sentence pattern where BVA is structurally supported.\footnote{What is unexplained is why SRC (which also lacks binder preceding bindee) is accepted so frequently, more or less on par with SOV. Note though that the parallel seems to be between SOV and SRC on the one hand, and OSV and ORC on the other, not between WCO and ORC, which the hypotheses of similarity between M-WCO and R-WCO would lead one to expect.}
WCO, on the other hand, is accepted much more rarely; indeed, most individuals (approximately 2/3) accepted BVA in ORC sentences at least sometimes, whereas only a small minority (approximately 1/5) ever accepted BVA in WCO sentences. 
Though this does not constitute a proof of difference between the two sentence types, it is certainly consistent with one, and serves as a meaningful ``sanity check'' on the analysis to be performed. 
That is, we are not going to be ``excavating'' a difference where one does not appear to exist superficially, but rather, showing that the already evident pattern can be shown to be categorical once sources of noise are removed.

In the analysis that follows, we find that the attentiveness-checking sub-experiments were not required in order to reveal a clear pattern of behavior. 
This situation has parallels in previous investigations of Japanese: \cite{Hoji2022c} reported similar results for at least some of the experiments performed. 
In essence, it seems that the DR and Coref tests are sufficiently difficult to ``pass'' if one is not paying attention/comprehending the intended questions that they eliminate both those with NFS effects and those who are not taking the experiment ``correctly''. 
We thus treat the sub-experiment questions as simply ``fillers'' for the sake of the analysis. \footnote{We have made our data available at Haruka Fukushima's git hub (\url{https://github.com/fuyunoharu}) those who are interested may verify that the addition of various sub-experiments does not make a qualitative difference in what is to be discussed, although it occasionally makes a (in our view non-substantive) quantitative one. }

We begin with the M-WCO case. 
Consider three possible classifications for a given informant:
\begin{exe}
    \ex BVA response patterns (matrix):
    \begin{xlist}
        \ex Accepting BVA in OSV at least sometimes, never accepting BVA in WCO.
        \ex Never accepting BVA in OSV or WCO
        \ex Accepting BVA in WCO at least sometimes
    \end{xlist}
\end{exe}

Those of type a above may be considered to have a clear structural contrast in BVA acceptability; even they do not accept BVA when structural conditions do not permit it, and do accept it when structural conditions do permit it, including at leats sometimes when binder does not precede bindee. 
Those of type b still consistently reject BVA when structural conditions do not permit it, but they do not show the clear contrast desired; they are consistent with structure-based BVA generalizations, but do not actively support them. 
Those of type c go against structure-based BVA generalizations. Our hypotheses predict (and as discussed above, previously results have shown) that, when proper noise control is applied, removing the effects of non-structural sources of BVA, the type-c cases should be entirely eliminated. 
This is indeed what we find with this dataset.
To eliminate such non-structural sources, we use both DR and Coref judgements. 
Specifically, if an individual consistently rejects DR and Coref in the WCO analogue sentences, and at least sometimes accept them in sentences where structural factors permit them (SOV and OSV sentences), we consider this a clear indication that non-structural factors are not available to support the DR/Coref reading in question. 
In particular, if DR, using the BVA binder as the thing to be ``distributed over'', follows a consistent structural pattern, the binder in question seems unable to participate in non-structural sources of readings for the participant in question, and likewise for the bindee and Coref, using the BVA bindee as the element that should ``corefer''. 
With the hypothesis the sources in question are the same for BVA (see discussion in the \cite{Plesniak2023}), we predict that BVA involving that binder and that bindee should not be available via non-structural sources, for the individual in question. 
The results of this analysis are shown below, where Y and N represent ``accepting at least sometimes with BVA'' and ``never accepting with BVA'' respectively:
\begin{table}[h]
    \centering
    \scalebox{0.9}{
    \begin{tabular}{l|l|l|l|l|l}
       BVA  & Passes Coref+DR tests & Passes DR test only & 
Passes Coref test only& Passes neither& total \\
    \hline       
OSV:Y WCO:N & 5&8&3&7&23\\
    \hline       
OSV:N WCO:N & 5&4&5&3&17\\
    \hline       
OSV:Y/N WCO:Y & 0&2&1&6&9\\
    \hline       
Total & 10&14&9&16&49\\
  \hline
    \end{tabular}
    }
     \caption{Matrix BVA judgement patterns, classified with reference to noise control}
    \label{tab:table2}
\end{table}

As can be seen from the table, most WCO-accepters pass neither diagnostic test, and none of them pass both. 
We can thus say that no individual for whom we are confident that their BVA judgements are based on a structural source is such that they accept BVA in WCO sentences. 
On the other hand, of the 10 individuals that do pass all the relevant diagnostics, none of them accept BVA in WCO, and half of them do show the clear OSV vs. WCO contrast, demonstrating the clear role of structural conditions in permitting BVA in matrix context, even in the absence binder-bindee precedence.

If we repeat this experiment in the relative clause context, substituting SRC for SOV/OSV and ORC for WCO, in both the BVA classification and the DR/Coref diagnostics, a very different result emerges:\footnote{We have examined other combinations of sentence patterns, such as adding SRC to SOV and OSV rather than substituting it. The basic results remain unchanged.
}
\begin{table}[h]
    \centering
    \scalebox{0.9}{
    \begin{tabular}{l|l|l|l|l|l}
       BVA  & Passes Coref+DR tests & Passes DR test only & 
Passes Coref test only& Passes neither& total \\
    \hline       
SRC:Y ORC:N & 0&1&4&7&12\\
    \hline       
SRC:N ORC:N & 0&0&1&5&6\\
    \hline       
SRC:Y/N ORC:Y &0&2&3&26&31\\
    \hline       
Total & 0&3&8&38&49\\
  \hline
    \end{tabular}
    }
    \label{tab:table3}
    \caption{RC BVA judgement patterns, classified with reference to noise control }
\end{table}

This pattern, in which no individual passes the noise control tests, is discussed at some length in \cite{Plesniak2022c}, and it is precisely what one might expect to see if ORC is not in fact ``equivalent to'' WCO. 
Namely, the DR and Coref diagnostics are predicated on the use of some sentence type as an exemplar of a case where the conditions for the structural source of BVA are not met. 
WCO is such a case, and as such, the diagnostics work just fine for the matrix case discussed above. 
However, if ORC is not such a case, i.e., the conditions for the structural source of BVA are met in ORC, then the substitution of ORC for WCO in the diagnostic procedures is illegitimate. 
It thus produces a diagnostic test which, rather than screening out those whose binders or bindees cannot participant in non-structural BVA, instead screens out individuals according to some arbitrary criterion, which is, in fact, not satisfied by anyone.
This is a sign that the hypothesis that ORC is structurally equivalent to WCO in the relevant sense is incorrect.
To further drive home this point, we can instead use the ``good'' diagnostic criteria from the matrix case (where the DR and Coref tests are based on SOV/OSV vs. WCO) and apply the results to the participants classified by their SRC vs. ORC judgements:
\begin{table}[h]
    \centering
    \scalebox{0.9}{
    \begin{tabular}{l|l|l|l|l|l}
       BVA  & Passes Coref+DR tests & Passes DR test only & 
Passes Coref test only& Passes neither& total \\
    \hline       
SRC:Y ORC:N & 3&3&4&2&12\\
    \hline       
SRC:N ORC:N & 2&1&1&2&6\\
    \hline       
SRC:Y/N ORC:Y & 5&10&4&12&31\\
    \hline       
Total & 10&14&9&16&49\\
  \hline
    \end{tabular}
    }
    \caption{RC BVA judgement patterns, using M BVA noise control }
    \label{tab:table4}
\end{table}

The diagnostics no longer ``eliminate'' everyone; we know from the matrix case already discussed that 10 individuals pass both tests. 
However, of those 10 individuals, half accept BVA in ORC constructions. 
It is thus not true at all that noise control reduces so-called R-WCO BVA acceptance to 0, in clear contrast with M-WCO BVA acceptance. \footnote{Again referring to our data available online, we can find no reasonable criterion or combination thereof whereby one can isolate R-WCO accepters from non-accepters, unlike M-WCO, where it is fairly easy to do so.}
As such, BVA in ORC constructions seems to be available regardless of whether or not an individual makes use of non-structural sources of BVA. 
This replicates the results of \cite{Fukushima2024}, without the confound of word order, providing cross-linguistic evidence for the differing status of relative and matrix clauses with respect to weak-crossover effects. \footnote{One may ask, if ORC meets the conditions for the structural source of BVA, can it be substituted for OSV/SVO, such that in DR/Coref diagnostics tests and BVA classification, we consider ORC vs. WCO? The answer is yes, and indeed, the results come out as predicted:
\scalebox{0.85}{
\begin{tabular}{l|l|l|l|l|l}
       BVA  & Passes Coref+DR tests & Passes DR test only & 
Passes Coref test only& Passes neither& total \\
    \hline       
ORC:Y WCO:N& 2&9&4&8&23\\
    \hline       
ORC:N WCO:N & 0&6&4&7&17\\
    \hline       
ORC:Y/N WCO:Y &0&2&1&6&9\\
    \hline       
Total & 2&17&9&21&49\\
  \hline
    \end{tabular}
}   
}
\section{Conclusion}
\label{sec:conclusion}
In this paper, we have shown evidence in favor of theories that treat relative clauses as structurally distinct from their matrix clause equivalents, at least with respect to the relevant factors for BVA interpretations. 
Crucially, we find no evidence for a ``relative weak-crossover effect'', while replicating previous results demonstrating the robust existence of a weak-crossover effect in a matrix clause. 
This result, taken together with that of \cite{Fukushima2024}, suggest that this matrix-relative clause contrast is cross-linguistically reproducible, including in languages with crucially differing word orders. 
While it goes without saying that much more investigation can and should be done, we think it is a fair conclusion that the onus is now more strongly on those who support the existence of an R-WCO effect to provide evidence for that claim. 
Until such evidence is forthcoming, theories that do not predict parallel behavior across the two types of clauses are a better match for the observed patterns of judgement, at least for speakers of Japanese and English.

This is not to say that there are no tendency-based contrasts one could establish. 
In this experiment, BVA was less frequently accepted in ORC sentences than in SVO ones. 
However, factors such as linear word order provide plausible explanations for this contrast; indeed, in \cite{Fukushima2024}'s English data, where the two were matched in word order, no such contrast emerged. 
Relative cognitive difficulty may also be a factor. 
Less easily explained is the observed contrast in frequency of acceptance of BVA in SRC and ORC sentences. 
The existence of such a contrast, if it can be reproduced, is surely of interest for future studies. 
However, we stress that this contrast does not seem to be based on categorical structural differences, unlike the contrast between OSV and WCO. 
As such, it cannot be termed a ``weak-crossover effect'' based on current evidence, and it may well be reducible to issues of cognitive difficulty and/or pragmatics, though we cannot at this time rule out a more significant explanation.

As a starting point for further addressing this and other questions relating to ORC sentences, we would like to mention one hypothesis put forth by Hajime Hoji (p.c., Fall 2023). 
Suppose that at least some cases of ORCs are derived not from canonical SOV-like structures, but from a more complex ones, such as a major subject construction. 
For example, the nominal in (\ref{23}) below might derive from a clause like (\ref{25}), rather than one like (\ref{24}).
\begin{exe}
    \ex \label{23} 
    \gll Soko-no	syatyoo-ga renrakusita Toyota-igai \\
 	it-GEN 	president-NOM contacted Toyota-others.than\\
  \trans `Others than Toyota that their president contacted'
    \ex \label{24} 
     \gll Soko-no	syatyoo-ga Toyota-igai-ni renrakusita \\
 	it-GEN president-NOM Toyota-others.than-DAT contacted \\
 	 \trans `Their president contacted others than Toyota'
    \ex \label{25} 
    \gll Toyota-igai-ga	soko-no	syatyoo-ga renrakusita\\
 	Toyota-others.than-NOM it-GEN president-NOM contacted\\
 	 \trans  `(As for) others than Toyota, their president contacted (them)'
\end{exe}

If this is indeed the case, then it may not be so much that R-WCO does not exist, but rather, that ORCs are ambiguous between two underlying structural parses, which are made homophonous due to the deletion of the crucially disambiguating element from the clause (since it serves as the head noun which the clause modifies).\footnote{Other structures, such as the ``Deep OS'' parse of~\cite{Ueyama1998}, would serve to create a similar effect.}

We are not currently in a position to test this hypothesis given the current data. 
However, as a preliminary attempt, we did include major subject (MS) sentence types for participants to judge. 
Interestingly, of the individuals in Table \ref{tab:table4} who pass both the DR and Coref tests, there are three who consistently reject BVA in MS constructions, and those individuals all also consistently reject BVA in ORCs.
This is certainly consistent with Hoji's hypothesis, namely that ORC BVA is covert MS BVA. 
However, it is far from conclusive at this point, and a dedicated experiment is required to tease apart these possibilities. 
As such, just as \cite{Fukushima2024} found a robust M-WCO vs. R-WCO contrast, but could not distinguish whether it was due to structural or word-order-based properties of these constructions, we have again found such a contrast and further, demonstrated it cannot be reduced to word-order-based properties, but could not distinguish exactly what structure or structures are involved. 
A third experiment is thus required to distinguish between ambiguity-based accounts, where R-WCO effects do exist but are marked by alternative parses, and one like \cite{Bekki2023}'s, where R-WCO effects are simply non-existent. 
Therefore, while we consider the structural nature of M-WCO-R-WCO contrast to be established, the nature of its ultimate interpretation remains an open question.

\subsubsection{\ackname}
This work was partially supported by Japan Science and Technology Agency (JST) CREST Grant Number JPMJCR20D2.22 and Japan Society for the Promotion of Science (JSPS) KAKENHI Grant Number JP18H03284.

%
\bibliography{fukushima}

\end{document}